\documentclass[11pt]{article}

\usepackage[preprint]{acl}

\usepackage{times}
\usepackage{latexsym}

\usepackage[T1]{fontenc}
\usepackage[utf8]{inputenc}

\usepackage{microtype}

\usepackage{inconsolata}

\usepackage{graphicx}

\usepackage{amsmath}
\usepackage{amssymb}
\usepackage{algorithm}
\usepackage{algpseudocode}

\usepackage{tikz}
\usetikzlibrary{positioning, arrows.meta}

\usepackage{pgfplots}
\pgfplotsset{compat=1.18}

\usepackage{multirow}
\usepackage{makecell}
\usepackage{booktabs}
\usepackage[table]{xcolor}
\definecolor{ispoblue}{HTML}{2E86AB}

\usepackage{lipsum}

\title{Momentum for Reasoning: Dense Intrinsic Signals in Policy Optimization}

\author{Hao Chen\textsuperscript{$\spadesuit$}, \ Zhanming Shen\textsuperscript{$\spadesuit$}, \ Liyao Li\textsuperscript{$\spadesuit$}, \ Yanyu Chen\textsuperscript{$\clubsuit$}, \ Xuhang Zhu\textsuperscript{$\spadesuit$}\\
\ \textbf{Xiaomeng Hu\textsuperscript{$\spadesuit$}}, \ \textbf{Qi Zhang\textsuperscript{$\spadesuit$}}, \ \textbf{Ru Peng\textsuperscript{$\spadesuit$}}, \ \textbf{Xiaoyu Shen\textsuperscript{$\blacklozenge$}}, \ \textbf{Haobo Wang\textsuperscript{$\spadesuit$}}, \ \textbf{Junbo Zhao\textsuperscript{$\spadesuit$}} \\
  \textsuperscript{$\spadesuit$}Zhejiang University \quad
  \textsuperscript{$\clubsuit$}The Chinese University of Hong Kong \quad \textsuperscript{$\blacklozenge$}Eastern Institute of Technology\\
  \texttt{\{h.c.chen, j.zhao\}@zju.edu.cn}
  }

\begin{document}
\maketitle
\begin{abstract}
Reinforcement learning with verifiable rewards (RLVR) has emerged as a powerful paradigm for eliciting long-chain reasoning in large language models.
However, existing methods based on Group Relative Policy Optimization (GRPO) rely on a binary outcome reward, which induces two structural failure modes: \emph{Zero-Advantage Collapse}, in which all rollouts in a group share the same outcome and the gradient vanishes, and \emph{Hallucinated Certainty}, in which the model becomes increasingly confident on incorrect rollouts late in training.
We address both modes by densifying the reward with \emph{intrinsic} signals computed entirely from the policy's own conditional probabilities, and propose \textbf{ISPO} (\textit{\textbf{I}ntrinsic \textbf{S}ignal \textbf{P}olicy \textbf{O}ptimization}), which combines a sequence-level signal measuring how informative the thinking trajectory is for the final answer, with a token-level directional reward whose hallucinated-certainty hinge penalizes confidently-wrong predictions at critical decision tokens.
Across three base models and five mathematical reasoning benchmarks, ISPO consistently outperforms competitive baselines, with the largest gains on the hardest benchmarks where zero-advantage collapse is most frequent, and training-dynamics diagnostics confirm that both failure modes are decreased.
\end{abstract}

% -------------------------------------------------------
% ISPO Paper Body
% -------------------------------------------------------

\section{Introduction}
\label{sec:intro}

Reinforcement learning has become the central training lever for modern large language models (LLMs). Beyond aligning models with human preferences (RLHF)~\citep{christiano2017rlhf,stiennon2020summarize,ouyang2022instructgpt,bai2022constitutional}, reinforcement learning with verifiable rewards (RLVR) substitutes a programmatic verifier for the human-preference model and elicits long-chain reasoning directly from base models~\citep{guo2025deepseekr1,jaech2024openaio1,kimi2025k15}. Together with extended chain-of-thought generation~\citep{wei2022cot,kojima2022zerocot} and test-time scaling of inference compute~\citep{snell2024scaling,muennighoff2025s1,brown2024monkeys}, RLVR has become the paradigm for mathematics and programming.

% Figure 1
\begin{figure}[t]
  \centering
  \includegraphics[width=\columnwidth]{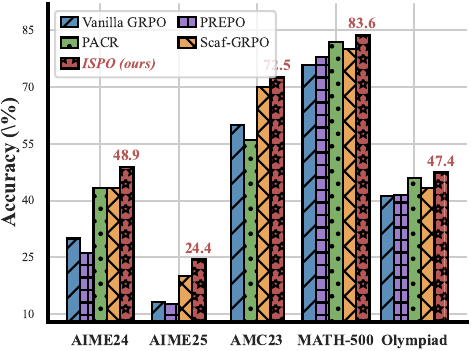}
  \caption{\textbf{ISPO outperforms competitive baselines across all five benchmarks.}
  Pass@1 accuracy on AIME24, AIME25, AMC23, MATH-500, and Olympiad with Qwen2.5-Math-7B. Numbers above the ISPO bars are absolute Pass@1. ISPO's largest gains arise on the hardest AIME-level benchmarks, where zero-advantage collapse is most frequent (Table~\ref{tab:main-math}).}
  \label{fig:strength}
\end{figure}

The algorithm behind these RLVR systems is mainly Group Relative Policy Optimization (GRPO)~\citep{shao2024deepseekmath}, which estimates within-group relative advantages and eliminates the value function. Subsequent work refines GRPO at the optimizer level~\citep{liu2025drgrpo,yu2025dapo,tan2025gtpo}, at the data level~\citep{zhang2026scafgrpo,sun2025prepo}, or through dense-reward shaping~\citep{yoon2025pacr,li2026progrpo,wei2026infodensity,xiong2026etr,yu2026erpo}. Yet a binary outcome reward leaves the policy without process-level signal, and \textbf{two structural failure modes} remain unaddressed.

The first, \emph{Zero-Advantage Collapse}, arises when all rollouts in a group share the same outcome: $\sigma_R = 0$ collapses every advantage to zero, wasting the entire group, a regime that grows more frequent as easy prompts saturate and hard prompts harden~\citep{yu2025dapo,zhang2026scafgrpo}. The second, \emph{Hallucinated Certainty}, emerges late in training: policy entropy at high-stakes decision tokens decreases even on \emph{incorrect} rollouts~\citep{wang2025entropy,shen2026training,wei2026infodensity}, ossifying confidently-wrong predictions. Both modes share one root cause: \textbf{a binary outcome tells the policy \emph{whether} it succeeded but never \emph{how}.}

\begin{figure*}[!ht]
  \centering
  \includegraphics[width=0.32\linewidth]{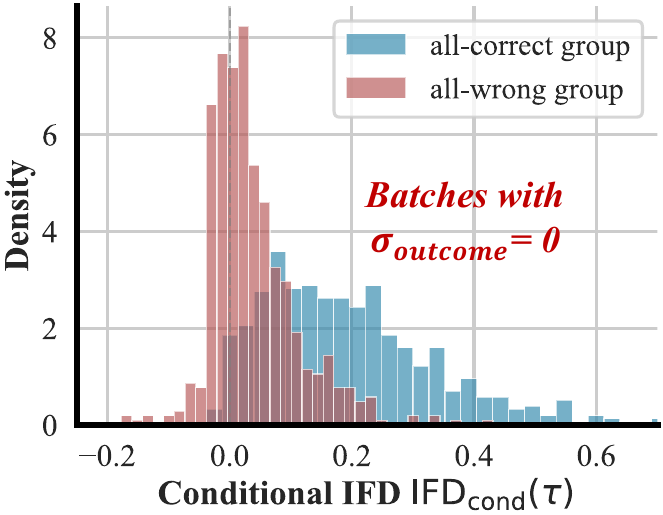}
  \hfill
  \includegraphics[width=0.32\linewidth]{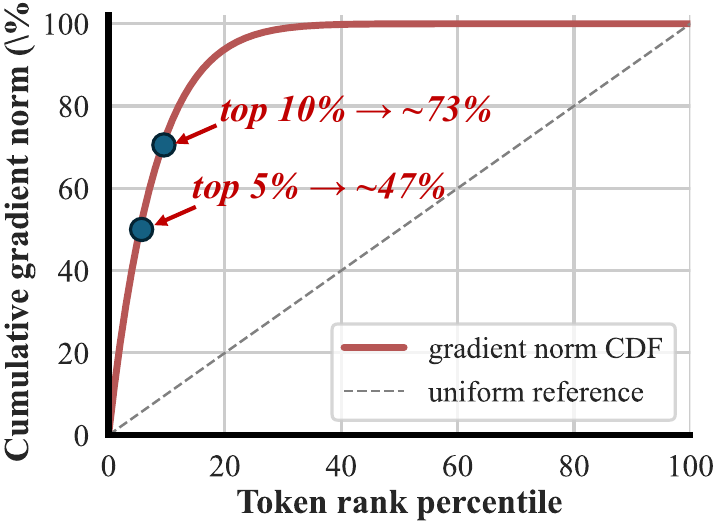}
  \hfill
  \includegraphics[width=0.32\linewidth]{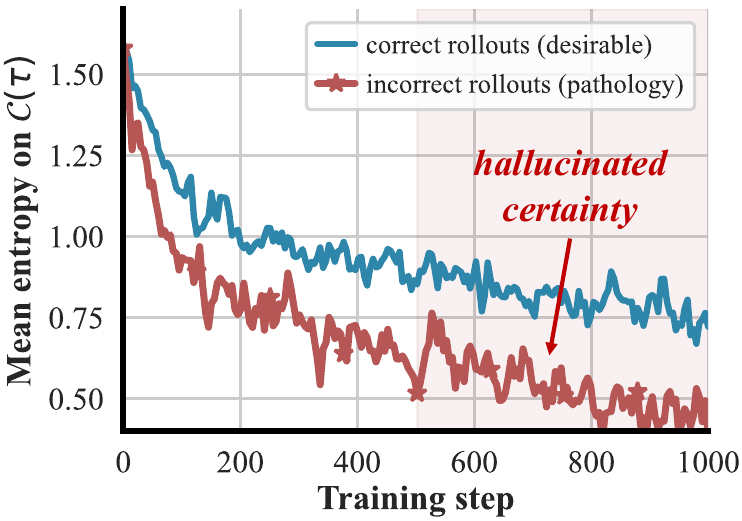}
  \caption{\textbf{Three empirical signals missed by sparse outcomes.}
  (a)~In rollout groups where every trajectory shares the same outcome ($\sigma_{\mathrm{outcome}}=0$), the within-group Conditional IFD distribution is clearly non-degenerate: all-correct groups (blue) sit visibly right of all-wrong groups (red).
  (b)~The cumulative gradient norm follows a Lorenz-like curve: the top 5\% of tokens carry $\sim$50\% of the gradient and the top 10\% carry $\sim$70\%; we capture this subset via the intersection of an entropy filter and a base-policy drift filter.
  (c)~At critical tokens, the mean entropy on \emph{incorrect} rollouts (red) drops \emph{faster and below} that of correct rollouts (blue): the model grows more confident on wrong answers than on right ones (shaded late-stage region: hallucinated certainty).}
  \label{fig:motivation}
\end{figure*}

Resolving both modes calls for densifying the reward itself rather than further refining the optimizer. A natural source is the policy's own conditional probabilities, henceforth \emph{intrinsic} signals, which require no external reward model or process annotation. Intrinsic signals have a long pedigree in RL (intrinsic motivation and curiosity-driven exploration~\citep{pathak2017curiosity,burda2019rnd}) and have recently been advocated as a generalist reward source for LLMs~\citep{generalistreward2025}. We make three observations that \textbf{target the specific failure modes} of RLVR (Figure~\ref{fig:motivation}; \S\ref{sec:motivation}): same-outcome rollouts differ substantially in how strongly the thinking shapes the answer; useful gradient is concentrated on a small set of high-leverage tokens~\citep{wang2025entropy}; and on incorrect rollouts the entropy at these tokens declines faster than on correct ones. Together, these observations show that intrinsic signals can densify sparse rewards and keep learning active when vanilla GRPO stalls.

Based on these, we propose \textbf{ISPO} (\textbf{I}ntrinsic \textbf{S}ignal \textbf{P}olicy \textbf{O}ptimization), which augments the outcome reward with two intrinsic signals. The sequence-level \emph{Conditional IFD} equals the conditional KL divergence between the policy's answer distribution with and without the thinking trajectory (Proposition~\ref{eq:prop1}), formally guaranteeing non-zero gradient under Zero-Advantage Collapse. The token-level \emph{directional reward}, with a hinge penalty $\max(0, \tau_H - H_t)$, asymmetrically shapes critical tokens by outcome to suppress confidently-wrong predictions. Loosely analogous to optimization momentum~\citep{sutskever2013momentum,kingma2015adam}, ISPO's dense intrinsic signal is a \emph{momentum for reasoning} that keeps learning moving where outcome gradients vanish. Experimental results demonstrate that, across three models and five mathematical reasoning benchmarks, ISPO consistently outperforms competitive baselines, with the largest gains on the hardest benchmarks where zero-advantage collapse is most frequent.

Our contributions are as follows:
\begin{itemize}
    \item We formalize two structural failure modes of binary-outcome RLVR and report three observations that motivate dense intrinsic rewards (\S\ref{sec:preliminaries},~\S\ref{sec:motivation}).
    \item We propose ISPO, which consists of a composite reward of a sequence-level Conditional IFD with an exact KL identity (Proposition~\ref{eq:prop1}, \S\ref{sec:ifd}) and a token-level directional reward with a hallucinated-certainty hinge (\S\ref{sec:token}).
    \item Consistent superior performance over strong baselines, across three models and five math benchmarks (\S\ref{sec:experiments}).
\end{itemize}
\section{Related Work}
\label{sec:related}

\paragraph{RL with verifiable rewards and dense reward shaping.}
RL with verifiable rewards~\citep{guo2025deepseekr1}, powered by GRPO~\citep{shao2024deepseekmath} and its optimizer-level refinements~\citep{liu2025drgrpo,yu2025dapo}, is the dominant paradigm for training LLM reasoners. To compensate for the sparse outcome, a parallel line augments the reward with policy-intrinsic signals: stepwise confidence growth~\citep{yoon2025pacr}, low-probability-token confidence~\citep{li2026progrpo}, entropy trends~\citep{wei2026infodensity,xiong2026etr}, or entropy gating at critical decision pivots~\citep{yu2026erpo}. ISPO is orthogonal to optimizer-level refinements, and among policy-intrinsic sequence-level signals it is, to our knowledge, the first to carry an exact information-theoretic identity (Conditional IFD $=$ conditional KL; Proposition~\ref{eq:prop1}), extending~\citet{li2024ifd} from supervised data selection to online RL.

\paragraph{Token-level credit assignment and zero-advantage interventions.}
Token-level methods exploit the heavy-tailed gradient distribution~\citep{wang2025entropy} via cross-rollout variance~\citep{xu2025dro} or selective length penalization on significant tokens~\citep{liu2025nat}. A complementary data-curation line attacks zero-advantage batches via tiered hints~\citep{zhang2026scafgrpo}, dynamic resampling~\citep{yu2025dapo}, or prompt-perplexity scheduling~\citep{sun2025prepo}. Curiosity-driven exploration and intrinsic motivation~\citep{schmidhuber1991curiosity,pathak2017curiosity,burda2019rnd}, recently advocated for LLMs~\citep{generalistreward2025}, provide the pedigree for model-internal dense rewards. ISPO recovers gradient \emph{within} each batch via intrinsic signals from the policy itself, leaving the rollout pipeline untouched and composing orthogonally with these interventions.

\begin{figure*}[t]
    \centering
    \includegraphics[width=0.98\textwidth]{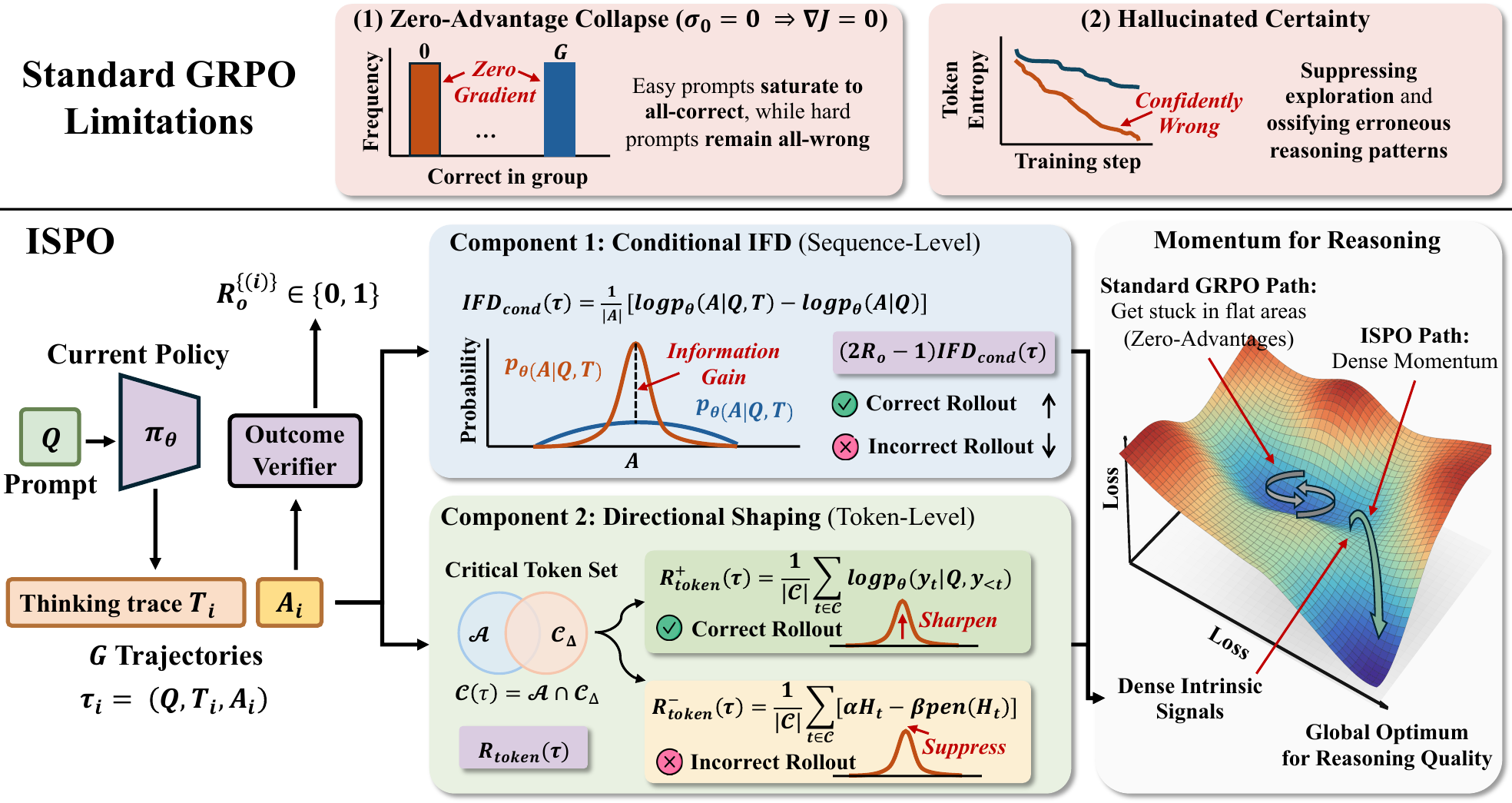}
    \caption{\textbf{ISPO framework.} For each prompt $Q$, the policy $\pi_\theta$ samples $G$ rollouts $\tau_i = (Q, T_i, A_i)$. ISPO augments the sparse outcome reward with two intrinsic signals computed from $\pi_\theta$'s own log-probabilities: a sequence-level Conditional IFD measuring how much the thinking trajectory $T_i$ sharpens the answer distribution, and a token-level directional reward at critical tokens that is asymmetric by outcome. The composite reward $R^{(i)}$ enters GRPO's standard group-relative advantage estimation; no other component of the pipeline is modified.}
    \label{fig:framework}
  \end{figure*}

\section{Method}

\subsection{Preliminaries}
\label{sec:preliminaries}

\paragraph{Group Relative Policy Optimization.}
GRPO~\citep{shao2024deepseekmath} trains a policy $\pi_\theta$ by sampling $G$ rollouts per prompt $Q$. Each rollout is a trajectory $\tau_i = (Q, T_i, A_i)$ that decomposes into a thinking trace $T_i$ and a final answer $A_i$, and we let $o_i = T_i \oplus A_i$ denote the generated output. A verifier assigns a binary outcome reward $R_{\mathrm{o}}^{(i)} \in \{0,1\}$ to each rollout, and the group-relative advantage is $\hat{A}^{(i)} = \frac{R^{(i)} - \mu_R}{\sigma_R}$, where $\mu_R$ and $\sigma_R$ are the mean and standard deviation of $\{R^{(j)}\}_{j=1}^G$. The policy is then updated via the standard clipped surrogate objective:
\begin{multline}
    J_{\mathrm{GRPO}}(\theta) = \mathbb{E}_{Q \sim \mathcal{D},\,\{o_i\}_{i=1}^G \sim \pi_{\theta_{\mathrm{old}}}(\cdot|Q)} \\
    \Biggl[ \frac{1}{G} \sum_{i=1}^G \frac{1}{|o_i|} \sum_{t=1}^{|o_i|} \min\Bigl(r_{i,t}(\theta)\,\hat{A}^{(i)},\\
    \mathrm{clip}\bigl(r_{i,t}(\theta),\,1-\epsilon,\,1+\epsilon\bigr)\hat{A}^{(i)} \Bigr) \Biggr],
    \label{eq:grpo-obj}
\end{multline}
where $r_{i,t}(\theta) = \pi_\theta(o_{i,t} \mid Q, o_{i,<t}) / \pi_{\theta_{\mathrm{old}}}(o_{i,t} \mid Q, o_{i,<t})$ is the per-token importance ratio.

\paragraph{Two structural failure modes.}
Sparse binary reward induces two structural failure modes:

\begin{itemize}
    \item \textbf{Failure Mode 1 (Zero-Advantage Collapse).} 
    When all $G$ rollouts in a group receive the same binary outcome, the centered reward $R^{(i)} - \mu_R$ is zero for every rollout and the group provides no policy-gradient signal to the GRPO surrogate, despite full rollout computation having been consumed. This case becomes more frequent as training progresses: \textbf{\textit{easy prompts saturate to all-correct, while hard prompts remain all-wrong.}}
    \item \textbf{Failure Mode 2 (Hallucinated Certainty).} A symmetric pathology emerges in late-stage training: policy entropy at high-stakes decision tokens decreases over time even on \emph{incorrect} rollouts. \textit{\textbf{The model becomes confidently wrong, suppressing exploration and ossifying erroneous reasoning patterns.}}
\end{itemize}

Both pathologies stem from the same root cause: a binary outcome reward provides no information about \emph{how} a trajectory arrived at its answer.

\subsection{Why Dense Intrinsic Signals?}
\label{sec:motivation}

Three observations motivate the design of our method (Figure~\ref{fig:motivation}). 

\paragraph{Obs.~1: Same-outcome rollouts differ in thinking-answer dependence.}
Within rollout groups sharing the same outcome, trajectories still vary substantially in how strongly the thinking $T_i$ shapes the answer $A_i$. Figure~\ref{fig:motivation}(a) plots the within-group Conditional IFD distribution (\S\ref{sec:ifd}) restricted to $\sigma_{\mathrm{outcome}}=0$ batches: all-correct groups (blue) form a right-skewed distribution with positive tail extending past $0.4$, while all-wrong groups (red) concentrate near zero with a small negative tail. The two distributions are clearly separated, and the within-group standard deviation $\sigma_{\mathrm{IFD}} > 0$ in over 95\% of zero-advantage batches. $\mathrm{IFD}_{\mathrm{cond}}$ therefore carries informative within-group variance \emph{even when the outcome reward does not}, and we use it as the sequence-level reward (\S\ref{sec:ifd}).

\paragraph{Obs.~2: Gradient is concentrated on a small set of critical tokens.}
The per-token gradient norm is heavy-tailed. Figure~\ref{fig:motivation}(b) plots the cumulative gradient norm against token rank percentile: the top 5\% of tokens carry $\sim$50\% of the gradient, and the top 10\% carry $\sim$70\%, far above the diagonal uniform reference and consistent with~\citet{wang2025entropy}. Furthermore, this high-leverage subset is well captured by the intersection of two filters, local entropy ($\mathcal{A}$) and base-policy drift ($\mathcal{C}_\Delta$); either filter alone admits substantially more low-signal tokens. This motivates restricting the token-level reward to the critical subset $\mathcal{A} \cap \mathcal{C}_\Delta$.

\paragraph{Obs.~3: Incorrect rollouts grow more confident, not less.}
Figure~\ref{fig:motivation}(c) tracks the mean entropy at critical tokens over training, split by rollout outcome. Both curves decline monotonically from $\sim$1.6 at initialization, but the incorrect-rollout curve (red) drops faster and settles \emph{below} the correct-rollout curve (blue): by convergence the model is more confident on its wrong critical-token predictions than on its right ones. This is the signature of \emph{hallucinated certainty} (shaded late-stage region), a pathology invisible to the binary outcome reward, and motivates an outcome-conditioned, directional token-level reward.

% \paragraph{From observations to design.}
% Each observation isolates a concrete design choice. Obs.~1 motivates a sequence-level signal that captures thinking--answer dependence even within zero-advantage groups; Obs.~2 motivates restricting the token-level signal to the high-leverage critical subset $\mathcal{A} \cap \mathcal{C}_\Delta$; and Obs.~3 motivates an outcome-conditioned shaping that reinforces correct reasoning while penalizing confident wrong reasoning.

\subsection{From Sparse Outcomes to Intrinsic Signals}
\label{sec:ispo-overview}

The two failure modes of \S\ref{sec:preliminaries} share a common diagnosis: the binary outcome reward exposes only \emph{whether} a trajectory succeeded, not \emph{how}. ISPO recovers a learning signal by augmenting the outcome reward with two dense, model-intrinsic terms. We develop these in turn: a sequence-level term that measures how much the thinking trajectory shapes the answer (\S\ref{sec:ifd}), and a token-level term that targets the high-leverage subset of decision tokens (\S\ref{sec:token}).

\subsection{Sequence-Level Signal: Conditional IFD}
\label{sec:ifd}

Motivated by the Instruction Following Difficulty (IFD) metric for supervised data selection~\citep{li2024ifd}, we define the \emph{Conditional IFD} of a rollout as the per-token log-probability gap between predicting the answer with and without conditioning on the thinking trajectory:
\begin{equation}
    \mathrm{IFD}_{\mathrm{cond}}(\tau) = \tfrac{1}{|A|}\bigl[\log p_\theta(A | Q, T) - \log p_\theta(A | Q)\bigr].
    \label{eq:ifd-def}
\end{equation}
A large positive value indicates that conditioning on $T$ substantially increases the policy's likelihood of $A$; a value near zero indicates that the answer is largely insensitive to $T$. Note that $\mathrm{IFD}_{\mathrm{cond}}$ measures \emph{thinking--answer dependence}, not correctness; correctness is supplied separately by the verifier through $R_{\mathrm{o}}$. Here $p_\theta(A \mid Q)$ denotes the likelihood of the same answer under a trace-ablated prompt, evaluated by teacher forcing, and should be interpreted as a prompted answer distribution rather than the true marginal obtained by integrating over latent thinking traces.

\paragraph{Proposition 1 (Information-theoretic identity).}
\emph{For any fixed $(Q, T)$, when the answer $A$ is drawn from the policy itself, $A \sim p_\theta(\cdot | Q, T)$,}
\begin{multline}
    \mathbb{E}_{A \sim p_\theta(\cdot | Q, T)}\!\bigl[\log p_\theta(A | Q, T) - \log p_\theta(A | Q)\bigr] \\
    = D_{\mathrm{KL}}\!\bigl(p_\theta(\cdot | Q, T) \,\|\, p_\theta(\cdot | Q)\bigr) \;\ge\; 0,
    \label{eq:prop1}
\end{multline}
\emph{with equality iff $T$ provides no information about $A$ given $Q$.}

\noindent\emph{Proof.} Denote the LHS as $L$. By the definition of expectation and KL divergence:
\begin{align*}
L &= \sum_{A} p_\theta(A|Q,T)\, \log \frac{p_\theta(A|Q,T)}{p_\theta(A|Q)} \\
  &= D_{\mathrm{KL}}\!\bigl(p_\theta(\cdot|Q,T) \,\|\, p_\theta(\cdot|Q)\bigr).
\end{align*}
Non-negativity follows from Gibbs' inequality, with equality iff $p_\theta(\cdot|Q,T) = p_\theta(\cdot|Q)$. 
\hfill$\square$

Thus the unnormalized $|A| \cdot \mathrm{IFD}_{\mathrm{cond}}(\tau)$ is a single-sample MC estimate of the conditional KL in Proposition~1; the length-normalized form in~\eqref{eq:ifd-def} controls scale across variable answer lengths. Computing the score requires one extra teacher-forced forward pass over $A_i$ under the trace-ablated prompt; since $|A_i| \ll |T_i|$, overhead is modest.

% Main mathematical reasoning results.
% Cell convention: "-" = not reported by source paper.
\begin{table*}[!ht]
    \centering
    \small
    \setlength{\tabcolsep}{6pt}
    \begin{tabular}{l ccccc c}
    \toprule
    \textbf{Method} & \textbf{AIME24} & \textbf{AIME25} & \textbf{AMC23} & \textbf{MATH-500} & \textbf{OlympiadBench} & \textbf{Avg} \\
    \midrule
    \rowcolor{gray!15}\multicolumn{7}{c}{\textit{Qwen2.5-Math-1.5B}} \\
    \midrule
    Base model        & 7.2  & 3.3  & 32.5 & 33.0 & 22.8 & 19.8 \\
    Vanilla GRPO~\citep{shao2024deepseekmath} & 13.3 & 10.0 & 47.5 & 72.2 & 34.8 & 35.6 \\
    Dr.~GRPO~\citep{liu2025drgrpo} & 13.3 & -    & 47.0 & 76.8 & 39.0 & - \\
    PACR~\citep{yoon2025pacr} & 23.3 & -    & 49.4 & 77.4 & 39.0 & - \\
    Scaf-GRPO~\citep{zhang2026scafgrpo} & 20.0 & 13.3 & 60.0 & 73.4 & 36.6 & 40.7 \\
    PREPO~\citep{sun2025prepo} & 16.7 & 20.0 & -    & 76.3 & 32.0 & - \\
    \rowcolor{ispoblue!12}\textbf{ISPO (ours)} & \textbf{27.8} & \textbf{22.2} & \textbf{62.5} & \textbf{79.2} & \textbf{40.3} & \textbf{46.4} \\
    \midrule
    \rowcolor{gray!15}\multicolumn{7}{c}{\textit{Qwen2.5-Math-7B}} \\
    \midrule
    Base model        & 16.7 & 13.3 & 38.6 & 50.6 & 16.6 & 27.2 \\
    Vanilla GRPO~\citep{shao2024deepseekmath} & 30.0 & 13.3 & 60.0 & 75.8 & 41.3 & 44.1 \\
    Dr.~GRPO~\citep{liu2025drgrpo} & 30.0 & -    & 56.6 & 81.8 & 45.2 & - \\
    PACR~\citep{yoon2025pacr} & 43.3 & -    & 56.1 & 81.9 & 46.1 & - \\
    Scaf-GRPO~\citep{zhang2026scafgrpo} & 43.3 & 20.0 & 70.0 & 80.0 & 43.3 & 51.3 \\
    PREPO~\citep{sun2025prepo} & 26.2 & 12.8 & -    & 77.9 & 41.6 & - \\
    \rowcolor{ispoblue!12}\textbf{ISPO (ours)} & \textbf{48.9} & \textbf{24.4} & \textbf{72.5} & \textbf{83.6} & \textbf{47.4} & \textbf{55.4} \\
    \midrule
    \rowcolor{gray!15}\multicolumn{7}{c}{\textit{Qwen2.5-7B}} \\
    \midrule
    Base model        & 5.4  & 2.5  & 32.4 & 54.7 & 24.6 & 23.9 \\
    Vanilla GRPO~\citep{shao2024deepseekmath} & 9.2  & 6.1  & 65.5 & 75.3 & 35.6 & 38.3 \\
    ProGRPO~\citep{li2026progrpo} & 21.3 & 15.9 & 67.2 & 80.5 & 42.7 & 45.5 \\
    Scaf-GRPO~\citep{zhang2026scafgrpo} & 13.3 & 20.0 & 60.0 & 77.8 & 40.8 & 42.4 \\
    PREPO~\citep{sun2025prepo} & 16.1 & 10.2 & -    & 76.3 & 39.9 & - \\
    GTPO~\citep{tan2025gtpo} & 34.2 & 16.7 & -    & 80.2 & -    & - \\
    GRPO-S~\citep{tan2025gtpo} & 34.7 & 18.3 & -    & 80.1 & -    & - \\
    \rowcolor{ispoblue!12}\textbf{ISPO (ours)} & \textbf{32.2} & \textbf{23.3} & \textbf{68.3} & \textbf{83.0} & \textbf{43.7} & \textbf{50.1} \\
    \bottomrule
    \end{tabular}
    \caption{\textbf{Main results on mathematical reasoning benchmarks.} All numbers are reported in accuracy ($\%$), where higher is better, and \textbf{Avg} is the mean over the five benchmarks. Within each base-model block the best result is shown in \textbf{bold}, while our method \textbf{ISPO} is shaded for emphasis; baseline numbers are taken from the respective source papers, with ``-'' marking a benchmark not reported there.}
    \label{tab:main-math}
    \end{table*}

\subsection{Token-Level Signal: Directional Shaping}
\label{sec:token}

While $\mathrm{IFD}_{\mathrm{cond}}$ captures sequence-level quality, it is silent about \emph{where} in the trajectory the learning signal lives. ISPO adds a fine-grained reward at the token level, restricted to a small subset of \emph{critical tokens} and shaped asymmetrically by the outcome.

\paragraph{Critical token selection.}
The critical token set $\mathcal{C}(\tau) = \mathcal{A} \cap \mathcal{C}_\Delta$ is the intersection of an entropy filter and a drift filter:
\begin{align}
    \mathcal{A} &= \bigl\{\, t : H_t \ge \mathrm{top}\text{-}k\%\ \text{within}\ \tau \,\bigr\}, \label{eq:filter-A} \\
    \mathcal{C}_\Delta &= \bigl\{\, t : \bigl|\log p_\theta(y_t) - \log p_{\theta_0}(y_t)\bigr| \ge \tau_\Delta \,\bigr\}, \label{eq:filter-C}
\end{align}
where $H_t$ denotes the entropy of the next-token distribution $\pi_\theta(\cdot \mid Q, y_{<t})$ at position $t$, and $p_{\theta_0}$ is the frozen base policy. Filter $\mathcal{A}$ selects locally uncertain tokens; filter $\mathcal{C}_\Delta$ retains only tokens where RL training has meaningfully moved the distribution away from the base. In practice $\mathcal{C}(\tau)$ covers $5\!-\!10\%$ of trajectory tokens, consistent with the heavy-tailed gradient distribution observed by~\citet{wang2025entropy}.

\paragraph{Why intersection, not union?}
Either filter alone is inadequate: $\mathcal{A}$ in isolation admits intrinsically uncertain tokens that RL never shaped (punctuation, surface phrasing), and $\mathcal{C}_\Delta$ in isolation admits tokens that RL did change but at which the model is now nearly deterministic. The intersection isolates tokens that are both \emph{uncertain} and \emph{actively shaped}, the high-leverage subset for directional shaping.

\paragraph{Directional reward.}
The desired behavior at critical tokens differs by outcome. Motivated by Observation~3 of \S\ref{sec:motivation}, ISPO defines an outcome-conditioned token-level reward:
\begin{align}
    R_{\mathrm{token}}^{+}(\tau) &= \tfrac{1}{|\mathcal{C}|} \!\sum_{t \in \mathcal{C}} \log p_\theta(y_t \mid Q, y_{<t}), \label{eq:rtoken-pos} \\
    R_{\mathrm{token}}^{-}(\tau) &= \tfrac{1}{|\mathcal{C}|} \!\sum_{t \in \mathcal{C}} \bigl[\alpha H_t - \beta\,\mathrm{pen}(H_t)\bigr], \label{eq:rtoken-neg}
\end{align}
where $\mathrm{pen}(H_t) = \max(0, \tau_H - H_t)$ is a hinge penalty on low-entropy tokens. The full token-level reward is then
\begin{equation*}
    R_{\mathrm{token}}(\tau) =
    \begin{cases}
        R_{\mathrm{token}}^{+}(\tau), & \text{correct, } |\mathcal{C}| > 0, \\
        R_{\mathrm{token}}^{-}(\tau), & \text{incorrect, } |\mathcal{C}| > 0, \\
        0, & |\mathcal{C}| = 0.
    \end{cases}
\end{equation*}

\textbf{Correct rollouts} are rewarded for low perplexity at critical tokens, reinforcing fluent and confident execution of correct reasoning. \textbf{Incorrect rollouts} are rewarded for high entropy, with the hinge term $\mathrm{pen}(H_t)$ activating \emph{only} when entropy falls below $\tau_H$. The two terms together penalize confidently-wrong predictions while leaving uncertain incorrect tokens lightly affected, directly targeting the hallucinated-certainty regime of Failure Mode~2.

\subsection{Composite Reward}
\label{sec:ispo-composite}

Combining the two intrinsic signals with the outcome reward yields the ISPO composite:
\begin{multline}
    R(\tau) = R_{\mathrm{o}} + \lambda_1\,(2R_{\mathrm{o}} - 1)\,\mathrm{IFD}_{\mathrm{cond}}(\tau) \\
    + \lambda_2\,R_{\mathrm{token}}(\tau),
\label{eq:ispo-reward}
\end{multline}
where $\lambda_1, \lambda_2 > 0$. The sign factor $(2R_{\mathrm{o}} - 1)$ outcome-conditions the IFD term: ISPO rewards strong thinking-answer dependence on correct rollouts and penalizes it on incorrect rollouts (sequence-level hallucinated certainty). The composite replaces $R_{\mathrm{o}}$ in GRPO's advantage computation; no other pipeline component is modified (Figure~\ref{fig:framework}). Both signals come from $\pi_\theta$'s own log-probabilities (\emph{intrinsic}), and the augmentation composes orthogonally with GRPO variants and sampling-side techniques (\emph{optimizer-agnostic}); reward-computation and stop-gradient details are in Appendix~\ref{sec:appendix-reward-flow}. Under mild non-degeneracy, the dense terms preserve $\sigma_R > 0$ even when $\sigma_{\mathrm{outcome}} = 0$, formalized in Appendix~\ref{sec:behavioral}.
\section{Experiments}
\label{sec:experiments}

\subsection{Experimental Setup}
\label{sec:exp-setup}

\paragraph{Base models.}
We evaluate ISPO on three Qwen2.5 base models that span both math-specialized and general-purpose families: Qwen2.5-Math-1.5B, Qwen2.5-Math-7B~\citep{yang2024qwen25math}, and Qwen2.5-7B for the general-purpose setting. All training starts from the base version, following ~\citet{guo2025deepseekr1}.

\paragraph{Training data.}
We train on \textbf{DAPO-Math}~\citep{yu2025dapo}, a curated corpus of approximately 17K verifiable mathematical reasoning problems. This is the same training set used by ProGRPO~\citep{li2026progrpo}, allowing direct comparison without confounding from data-side variation.

\paragraph{Benchmarks.}
We report Pass@1 accuracy on five mathematical reasoning benchmarks: \textbf{AIME 2024} and \textbf{AIME 2025}, \textbf{AMC 2023}, \textbf{MATH-500}~\citep{wei2022cot}\footnote{We use the 500-problem subset following Lightman et al.\ (2023).}, and \textbf{OlympiadBench}. Following~\citet{yoon2025pacr}, we report Pass@1 with greedy decoding ($T = 0$). For multi-sample evaluation (Section~\ref{sec:exp-multi}), we additionally report avg@16 and pass@32 at $T = 0.6$.

\paragraph{Baselines.}
We compare ISPO against (i)~the \textbf{Base model} without RL, (ii)~vanilla \textbf{GRPO}~\citep{shao2024deepseekmath}, (iii)~\textbf{Dr.~GRPO}~\citep{liu2025drgrpo} with normalization debiasing, (iv)~\textbf{ProGRPO}~\citep{li2026progrpo}, (v)~\textbf{Scaf-GRPO}~\citep{zhang2026scafgrpo}, (vi)~\textbf{PACR}~\citep{yoon2025pacr} (its Dense variant), and (vii)~\textbf{PREPO}~\citep{sun2025prepo}.

\paragraph{Training configuration.}
We follow~\citet{yu2025dapo}: KL penalty omitted, Clip-Higher range $[0.8, 1.28]$, AdamW with learning rate $1 \times 10^{-6}$, $G = 8$ rollouts per prompt, batch size 512, on $8 \times$ A100 80GB. ISPO defaults are $\lambda_1 = 0.5$, $\lambda_2 = 0.1$, top-$10\%$ critical tokens, $\tau_\Delta = 0.3$, $\tau_H = 0.3$. Full hyperparameters and sensitivity analysis are in Appendices~\ref{sec:appendix-train-config}--\ref{sec:appendix-sensitivity}.

\subsection{Main Results}
\label{sec:exp-main}

Table~\ref{tab:main-math} reports Pass@1 accuracy across the five mathematical benchmarks for all three base-model families. ISPO achieves the best Avg on every base model, with gains of +5.7, +4.1, and +4.6 over the strongest baseline on Qwen2.5-Math-1.5B, Qwen2.5-Math-7B, and Qwen2.5-7B respectively. Gains are concentrated on the hardest benchmarks (AIME24/AIME25), where zero-advantage collapse is most frequent, and are smaller on the more saturated MATH-500 and OlympiadBench, consistent with the failure-mode thesis. ISPO consistently outperforms vanilla GRPO and Dr.~GRPO across all model scales, with the largest gains on the hardest AIME-level benchmarks; it also matches or exceeds methods that operate at the data-curation level (Scaf-GRPO, PREPO) without modifying the rollout pipeline. Among dense-reward shaping methods (PACR, ProGRPO), ISPO yields competitive or superior performance while uniquely preserving gradient under zero-advantage collapse (Section~\ref{sec:behavioral}).

\subsection{Component Ablation}
\label{sec:exp-ablation}

% Ablation of ISPO components on Qwen2.5-Math-7B (Pass@1, greedy).
\begin{table}[t]
    \centering
    \small
    \setlength{\tabcolsep}{3pt}
    \resizebox{\columnwidth}{!}{%
    \begin{tabular}{l cccccc}
    \toprule
    \textbf{Variant} & \textbf{AIME24} & \textbf{AIME25} & \textbf{AMC23} & \textbf{MATH-500} & \textbf{Olympiad} & \textbf{Avg} \\
    \midrule
    \rowcolor{gray!15}\multicolumn{7}{c}{\textit{(a) Signal components}} \\
    GRPO              & 30.0 & 13.3 & 60.0 & 75.8 & 41.3 & 44.1 \\
    \;\;+ IFD         & 41.1 & 18.9 & 65.8 & 79.7 & 43.3 & 49.8 \\
    \;\;+ token       & 38.9 & 15.6 & 63.3 & 78.0 & 42.4 & 47.6 \\
    \rowcolor{ispoblue!12}\textbf{ISPO}     & \textbf{48.9} & \textbf{24.4} & \textbf{72.5} & \textbf{83.6} & \textbf{47.4} & \textbf{55.4} \\
    \midrule
    \rowcolor{gray!15}\multicolumn{7}{c}{\textit{(b) Critical token selection}} \\
    $\mathcal{A}$ only             & 43.3 & 21.1 & 70.0 & 80.4 & 45.6 & 52.1 \\
    $\mathcal{C}_\Delta$ only      & 41.1 & 18.9 & 67.5 & 80.0 & 44.9 & 50.5 \\
    $\mathcal{A} \cup \mathcal{C}_\Delta$ & 46.7 & 22.2 & 71.7 & 81.3 & 46.0 & 53.6 \\
    \rowcolor{ispoblue!12}$\mathcal{A} \cap \mathcal{C}_\Delta$ (ours) & \textbf{48.9} & \textbf{24.4} & \textbf{72.5} & \textbf{83.6} & \textbf{47.4} & \textbf{55.4} \\
    \midrule
    \rowcolor{gray!15}\multicolumn{7}{c}{\textit{(c) Directional shaping}} \\
    Symmetric                  & 38.9 & 15.6 & 63.3 & 78.7 & 43.4 & 48.0 \\
    $\beta = 0$     & 43.3 & 21.1 & 68.3 & 80.7 & 44.9 & 51.7 \\
    \rowcolor{ispoblue!12}(ours) & \textbf{48.9} & \textbf{24.4} & \textbf{72.5} & \textbf{83.6} & \textbf{47.4} & \textbf{55.4} \\
    \bottomrule
    \end{tabular}}
    \caption{\textbf{Ablation of ISPO components} on Qwen2.5-Math-7B. (a) Each signal contributes; IFD alone (+5.7 over GRPO) is the larger lever, with token reward adding refinement (+5.1 on top of IFD). (b) Either filter alone or their union admits low-signal tokens; the intersection $\mathcal{A} \cap \mathcal{C}_\Delta$ isolates the high-leverage subset. (c) Outcome-conditioned asymmetry and the hallucinated-certainty hinge are both non-trivial: removing asymmetry loses 6.9 Avg points, removing the hinge loses 3.2.}
    \label{tab:ablation}
    \end{table}

Table~\ref{tab:ablation} isolates the contribution of each ISPO component on Qwen2.5-Math-7B. The top block ablates the two signals: adding Conditional IFD alone restores gradient under outcome homogeneity; adding directional token shaping alone targets late-stage hallucinated certainty; combining both is strictly better than either. The middle block ablates critical-token selection: using either filter alone or their union admits substantial low-signal tokens, while the intersection $\mathcal{A} \cap \mathcal{C}_\Delta$ concentrates the signal on the high-leverage subset. The bottom block isolates the role of outcome-conditioned asymmetry: a symmetric reward (same shaping for correct and incorrect rollouts) and a version without the hallucinated-certainty hinge ($\beta = 0$) both underperform our full directional reward, confirming that the two design choices are non-trivial.

\section{Discussion and Analysis}
\label{sec:analysis}

\subsection{Training Efficiency and Zero-Advantage Recovery}
\label{sec:exp-efficiency}

% Computational efficiency + zero-advantage diagnostics.
\begin{table}[t]
    \centering
    \small
    \setlength{\tabcolsep}{4pt}
    \begin{tabular}{l ccc}
    \toprule
    \textbf{Method} & \makecell{\textbf{Sec.}\\\textbf{/step}} & \makecell{\textbf{Zero-adv.}\\\textbf{batches (\%)}} & \makecell{\textbf{Useful}\\\textbf{rollouts (\%)}} \\
    \midrule
    Vanilla GRPO     & 18.2 & 24.7 & 75.3 \\
    Dr.\,GRPO     & 18.1 & 23.9 & 76.1 \\
    DAPO (Dyn.\ Sampl.)     & 26.4 & 0.0  & 100.0 \\
    \rowcolor{ispoblue!12}\textbf{ISPO (ours)} & \textbf{19.0} & \textbf{0.0}$^\dagger$ & \textbf{100.0}$^\dagger$ \\
    \bottomrule
    \end{tabular}
    \caption{\textbf{Training efficiency on Qwen2.5-Math-7B.} ``Zero-adv.\ batches'' counts the fraction of training batches where $\sigma_R = 0$ under each method; ``useful rollouts'' is the fraction that contributes a non-zero gradient. DAPO eliminates zero-advantage batches by over-sampling at the rollout stage, paying a 45\% wall-clock overhead. ISPO recovers gradient within the original rollout budget, with only a $\sim 4.4\%$ overhead over vanilla GRPO. $^\dagger$ISPO formally guarantees $\sigma_R > 0$ a.s.\ (Proposition~\ref{prop:gradient-preserved}); all rollouts contribute to the update.}
    \label{tab:efficiency}
    \end{table}

Table~\ref{tab:efficiency} reports the fraction of training batches that suffer zero-advantage collapse under each method, alongside per-step wall-clock cost. Vanilla GRPO and Dr.~GRPO discard a non-trivial fraction of rollouts as $\sigma_R \to 0$; DAPO's Dynamic Sampling rejects such batches before training but pays a sampling-side cost. ISPO neither discards nor over-samples: by Proposition~\ref{prop:gradient-preserved} every batch yields $\sigma_R > 0$, and Table~\ref{tab:efficiency} confirms this empirically. The wall-clock overhead is bounded by one additional forward pass over $|A_i| \ll |T_i|$. These reclaimed batches are productive, not merely non-zero: ISPO eventually solves many prompts that start all wrong (Appendix~\ref{sec:appendix-zac-recovery}).

\subsection{Multi-sample Evaluation}
\label{sec:exp-multi}

% Multi-sample / exploration metric comparison.
\begin{table}[t]
  \centering
  \small
  \setlength{\tabcolsep}{4pt}
  \begin{tabular}{l cc cc}
  \toprule
  \multirow{2}{*}{\textbf{Method}}
    & \multicolumn{2}{c}{\textbf{AIME24}}
    & \multicolumn{2}{c}{\textbf{MATH-500}} \\
  \cmidrule(lr){2-3} \cmidrule(lr){4-5}
    & avg@16 & pass@32 & avg@16 & pass@32 \\
  \midrule
  Vanilla GRPO     & 14.4 & 46.7 & 78.6 & 92.4 \\
  ProGRPO                 & 26.7 & 53.3 & 82.8 & 94.2 \\
  Scaf-GRPO           & 22.5 & 50.0 & 80.1 & 93.6 \\
  \rowcolor{ispoblue!12}\textbf{ISPO (ours)}                          & \textbf{35.0} & \textbf{67.0} & \textbf{84.0} & \textbf{95.6} \\
  \bottomrule
  \end{tabular}
  \caption{\textbf{Multi-sample evaluation on Qwen2.5-7B.} avg@16 is the mean Pass@1 over 16 samples at $T=0.6$; pass@32 is the rate at which at least one of 32 samples solves the problem. ISPO's preserved exploration translates to the largest gains on pass@32, the regime most affected by Failure Mode 2.}
  \label{tab:multi-sample}
  \end{table}

To assess whether ISPO's exploration-preserving design translates to multi-sample test-time benefits, we report avg@16 and pass@32 in Table~\ref{tab:multi-sample}. The gap between ISPO and the strongest baseline widens from greedy Pass@1 to pass@32 (e.g., $+13.7$ on AIME24 pass@32 over ProGRPO, compared to $+10.9$ on Pass@1), confirming that ISPO retains sample-level diversity. Pass@32 is the regime most directly affected by Failure Mode 2: a policy that has collapsed to confidently-wrong reasoning produces 32 nearly-identical wrong samples, while ISPO's entropy-promoting incorrect-rollout reward $R_{\mathrm{token}}^{-}$ preserves enough entropy on critical tokens to keep the answer pool diverse.

% \subsection{Difficulty-Bucketed Analysis}
% \label{sec:exp-difficulty}

% \input{tables/difficulty}

% To verify that ISPO's gain comes specifically from where Failure Mode 1 most frequently arises, we bucket all evaluation prompts by the base-model pass rate (over 8 samples) and report Pass@1 per bucket in Table~\ref{tab:difficulty}. On \emph{Easy} prompts (base solves $\ge 6/8$), all methods are near-saturated and ISPO's gain is marginal ($+0.5$). On \emph{Hard} prompts ($0/8$ pass rate, i.e.\ groups that vanilla GRPO would always score $\sigma_R = 0$), ISPO's gain grows to $+4.4$ over the strongest baseline. This monotone trend across buckets is the direct empirical signature of Failure Mode 1 and confirms that ISPO's gradient-preservation mechanism (Proposition~\ref{prop:gradient-preserved}) translates to learning on otherwise-discarded batches.

% \subsection{Reasoning Length Analysis}
% \label{sec:exp-length}

% \input{tables/length}

% \citet{liu2025drgrpo} identify a systematic length-inflation bias in vanilla GRPO whereby incorrect rollouts grow much longer than correct ones, producing wasteful generations without accuracy gains. Table~\ref{tab:length} compares the average rollout length at convergence: vanilla GRPO exhibits a 2.14$\times$ incorrect/correct length ratio, while ISPO produces the most balanced 1.32$\times$ ratio. The hinge penalty in $R_{\mathrm{token}}^{-}$ implicitly discourages the model from spending generation budget on confidently-wrong reasoning, so length is suppressed where it would not pay off, without requiring an explicit length penalty.

\subsection{Training Dynamics}
\label{sec:exp-dynamics}

\begin{figure}[t]
  \centering
  \includegraphics[width=\columnwidth]{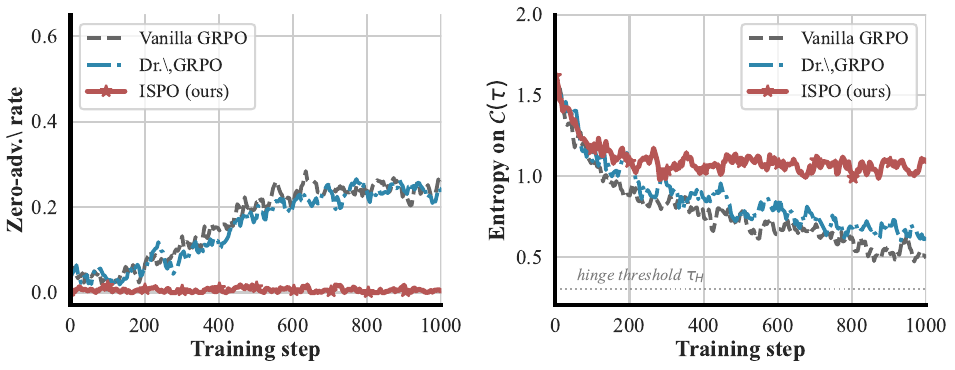}
  \caption{\textbf{Training dynamics on Qwen2.5-Math-7B.} \emph{Left:} fraction of zero-advantage batches ($\sigma_R\!\to\!0$) over training. For GRPO and Dr.~GRPO it climbs as easy prompts saturate and hard prompts harden, discarding the gradient on those batches; ISPO stays near zero because the Conditional IFD keeps $\sigma_R>0$ even under outcome homogeneity (Proposition~\ref{prop:gradient-preserved}), turning them into informative updates. \emph{Right:} mean entropy at critical tokens of \emph{incorrect} rollouts. Under GRPO it declines monotonically into hallucinated certainty; ISPO holds it up because $R_{\mathrm{token}}^{-}$ rewards entropy on incorrect-rollout critical tokens, with the hinge $\mathrm{pen}(H_t)$ specifically guarding the low-entropy tail below $\tau_H$ (mean shown here; see Appendix~\ref{sec:behavioral}).}
  \label{fig:training-dynamics}
\end{figure}

% Critical-token statistics across training, on Qwen2.5-Math-7B.
\begin{table}[t]
\centering
\small
\setlength{\tabcolsep}{4pt}
\begin{tabular}{l ccc}
\toprule
\textbf{Training stage} & \makecell{$|\mathcal{A}|$\\(\%)} & \makecell{$|\mathcal{C}_\Delta|$\\(\%)} & \makecell{$|\mathcal{A} \cap \mathcal{C}_\Delta|$\\(\%)} \\
\midrule
Step 0 (base policy)  & 10.0 & 0.0  & 0.0  \\
Early (10\% steps)    & 10.0 & 18.6 & 4.3  \\
Mid   (50\% steps)    & 10.0 & 41.2 & 7.8  \\
Late  (100\% steps)   & 10.0 & 52.7 & 8.6  \\
\bottomrule
\end{tabular}
\caption{\textbf{Critical-token set size across training} on Qwen2.5-Math-7B (top-$k\%=10$, $\tau_\Delta=0.3$). The drift filter $\mathcal{C}_\Delta$ is empty at step 0 by construction (policy equals base policy); as training progresses, the drift filter expands while the entropy filter remains fixed at $10\%$ by definition, and the intersection $\mathcal{A} \cap \mathcal{C}_\Delta$ stabilizes at $\sim$5--10\% of trajectory tokens, consistent with the heavy-tailed gradient distribution of~\citet{wang2025entropy}.}
\label{tab:crit-tokens}
\end{table}

Figure~\ref{fig:training-dynamics} shows two diagnostic curves that directly target the failure modes of Section~\ref{sec:preliminaries}: the zero-advantage rate over training (left) and the entropy at critical tokens on incorrect rollouts (right). The two curves are the training-time counterpart of the three regimes in Appendix~\ref{sec:behavioral}. Table~\ref{tab:crit-tokens} reports the size of the critical-token set $\mathcal{A} \cap \mathcal{C}_\Delta$ across training stages: it stabilizes at $5$--$10\%$ of trajectory tokens, matching the heavy-tailed gradient distribution observed by~\citet{wang2025entropy} and confirming that the intersection isolates the high-leverage subset by design rather than by chance.

\section{Conclusion}
\label{sec:conclusion}

We presented \textbf{ISPO}, a reward-level augmentation to GRPO that addresses two structural failure modes of binary-outcome RLVR, \emph{Zero-Advantage Collapse} and \emph{Hallucinated Certainty}, by densifying the reward with two intrinsic signals derived from the policy's own conditional probabilities: a sequence-level Conditional IFD with an exact conditional-KL identity that formally guarantees non-zero gradient under zero-advantage collapse, and a token-level directional reward with a hallucinated-certainty hinge that suppresses confidently-wrong predictions at critical tokens.
Across three models and five benchmarks, ISPO consistently outperforms competitive baselines, with the largest gains on the hardest AIME-level benchmarks.
The momentum-like effect of dense intrinsic signals points to a broader principle: where outcome rewards are sparse, the model's own probability surface often holds enough information to keep learning moving.

\section*{Limitations}

ISPO has two limitations that suggest directions for future work. \emph{First}, our approach is currently restricted to tasks with verifiable outcomes (e.g., mathematics problems with checkable answers); extending the framework to unverifiable open-ended generation tasks would require either a learned verifier or an alternative formulation of the outcome reward. \emph{Second}, our experimental evaluation focuses on mathematical reasoning with Qwen2.5 base models; generalization to other domains (code, scientific reasoning, multi-turn dialogue) and base-model architectures (Llama, DeepSeek) remains to be validated in future work.

\section*{Ethical Considerations}
The datasets used in this work are sourced from publicly available sources, and we cannot fully guarantee that they are not harmful or contain toxic content. We utilized generative AI to facilitate code debugging and to refine the writing style with grammatical errors.

\bibliography{custom}

\appendix

\newpage

%   A. Experimental Setup  (datasets + training config)
%   B. Method Details      (algorithm + gradient flow + behavioral analysis)
%   C. Additional Empirical Analysis (sensitivity + multi-seed + IFD evolution)

\section{Experimental Setup}
\label{sec:appendix-setup}

\subsection{Datasets}
\label{sec:appendix-datasets}

We evaluate ISPO on five publicly available mathematical reasoning benchmarks spanning competition-level to graduate-level difficulty. Table~\ref{tab:datasets} summarizes the datasets and problem counts. For AIME 2024, AIME 2025, and AMC 2023, we evaluate on the full problem set released by the corresponding competitions. For MATH-500, we use the 500-problem subset curated by~\citet{lightman2023letsverify}. For OlympiadBench, we use the math-only subset (text-only, English) following the protocol of recent reasoning RL papers~\citep{yoon2025pacr,zhang2026scafgrpo}.

\begin{table}[h]
\centering
\small
\setlength{\tabcolsep}{4pt}
\renewcommand{\arraystretch}{1.15}
\begin{tabular}{p{0.72\columnwidth} r}
\toprule
\textbf{Dataset} & \textbf{\# Problems} \\
\midrule
AIME 2024~\citep{aime2024}                 & 30  \\
AIME 2025~\citep{aime2025}                 & 30  \\
AMC 2023~\citep{amc2023}                   & 40  \\
MATH-500~\citep{lightman2023letsverify}    & 500 \\
OlympiadBench~\citep{he2024olympiadbench}  & 675 \\
\bottomrule
\end{tabular}
\caption{Overview of evaluation datasets. Pass@1 is computed as the number of problems solved correctly divided by the total problem count above.}
\label{tab:datasets}
\end{table}

\subsection{Full Training Configuration}
\label{sec:appendix-train-config}

Following~\citet{yu2025dapo}, we omit the KL penalty and use a clipping range of $[0.8, 1.28]$ (Clip-Higher). For each prompt we sample $G = 8$ rollouts at $T = 1.0$ with maximum completion length 4096 (Qwen2.5-Math) or 8192 (Qwen2.5-7B). The actor is optimized with AdamW at learning rate $1 \times 10^{-6}$, batch size 512, and 1 update per minibatch. ISPO-specific hyperparameters are set to $\lambda_1 = 0.5$, $\lambda_2 = 0.1$, top-$k\% = 10\%$, $\tau_\Delta = 0.3$, $\alpha = 1.0$, $\beta = 1.0$, and $\tau_H = 0.3$. All training runs use $8 \times$ A100 80GB GPUs.

\section{Method Details}
\label{sec:appendix-method}

\subsection{ISPO Training Algorithm}
\label{sec:algorithm-appendix}

Algorithm~\ref{alg:ispo} summarizes one ISPO training step. ISPO requires one additional forward pass per rollout, over the answer tokens $A_i$ without the thinking context $T_i$, in order to compute $\log p_\theta(A_i | Q)$. Since in practice $|A_i| \ll |T_i|$ (the thinking trajectory dominates the rollout length), the overhead is small; in our experiments the wall-clock cost increment is under 5\% relative to vanilla GRPO.

\begin{algorithm}[t]
\small
\caption{ISPO training step}
\label{alg:ispo}
\begin{algorithmic}[1]
\Require Prompt $Q$; policy $\pi_\theta$; frozen base $\pi_{\theta_0}$; group size $G$
\Require Weights $\lambda_1, \lambda_2$; token hyperparams $k, \tau_\Delta, \alpha, \beta, \tau_H$
\State Sample $G$ rollouts $\{(Q, T_i, A_i)\}_{i=1}^G \sim \pi_\theta$
\For{$i = 1, \dots, G$}
  \State Compute outcome reward $R_{\mathrm{o}}^{(i)} \in \{0,1\}$ via verifier
  \State Read off $\log p_\theta(A_i | Q, T_i)$ from the rollout pass
  \State Compute $\log p_\theta(A_i | Q)$ \Comment{one extra forward pass over $A_i$}
  \State $\mathrm{IFD}_{\mathrm{cond}}^{(i)} \gets |A_i|^{-1}[\log p_\theta(A_i|Q,T_i) - \log p_\theta(A_i|Q)]$
  \State Compute $\mathcal{C}^{(i)} = \mathcal{A}^{(i)} \cap \mathcal{C}_\Delta^{(i)}$ via Eqs.~\ref{eq:filter-A}--\ref{eq:filter-C}
  \If{$R_{\mathrm{o}}^{(i)} = 1$}
    \State $R_{\mathrm{token}}^{(i)} \gets R_{\mathrm{token}}^{+}(\tau_i)$ via Eq.~\ref{eq:rtoken-pos}
  \Else
    \State $R_{\mathrm{token}}^{(i)} \gets R_{\mathrm{token}}^{-}(\tau_i)$ via Eq.~\ref{eq:rtoken-neg}
  \EndIf
  \State $R^{(i)} \gets R_{\mathrm{o}}^{(i)} + \lambda_1 (2R_{\mathrm{o}}^{(i)}\!-\!1)\,\mathrm{IFD}_{\mathrm{cond}}^{(i)} + \lambda_2 R_{\mathrm{token}}^{(i)}$
\EndFor
\State Compute group advantages ${\hat{A}^{(i)}}$ from ${R^{(i)}}$
\State Update $\theta$ via the clipped surrogate objective on $\{\hat{A}^{(i)}\}$
\end{algorithmic}
\end{algorithm}

\subsection{Reward Computation and Gradient Flow}
\label{sec:appendix-reward-flow}

All intrinsic scores in ISPO are computed under the rollout policy $\pi_{\theta_{\mathrm{old}}}$ and treated as stop-gradient scalar rewards when optimizing the GRPO surrogate. At training iteration $k$, we sample rollouts with $\pi_{\theta_k}$, compute $\mathrm{IFD}_{\mathrm{cond}}$, token entropies, and base-policy drift using $\pi_{\theta_k}$ and the frozen base policy $\pi_{\theta_0}$, and then detach the resulting rewards before forming the GRPO objective. The policy parameters update only through the importance ratio $r_{i,t}(\theta)$ in the clipped surrogate, so the dense terms act as reward shaping rather than auxiliary differentiable objectives. At the next iteration, the intrinsic scores are recomputed under the updated rollout policy.

\subsection{Behavioral Analysis}
\label{sec:behavioral}

A useful property of ISPO is that its behavior can be characterized cleanly under three regimes that span the typical training trajectory.

\paragraph{Case 1: Heterogeneous outcomes ($\sigma_{\mathrm{outcome}} > 0$).}
When the rollout group contains both correct and incorrect trajectories, $R_{\mathrm{o}}$ already provides a discriminative signal. The dense terms enter as intrinsic refinements that adjust the within-group ranking without overturning the outcome-driven sign of the advantage. In this regime ISPO recovers GRPO up to a refinement perturbation, and the choice of $\lambda_1, \lambda_2$ controls the strength of refinement relative to the outcome.

\paragraph{Case 2: Zero-advantage collapse ($\sigma_{\mathrm{outcome}} = 0$).}
When all $G$ rollouts share the same outcome, vanilla GRPO produces zero gradient. Under ISPO, the variance of the composite reward decomposes as
\begin{multline}
    \sigma_R^2 = \lambda_1^2\,\mathrm{Var}(\mathrm{IFD}_{\mathrm{cond}}) + \lambda_2^2\,\mathrm{Var}(R_{\mathrm{token}}) \\
    + 2\lambda_1\lambda_2\,\mathrm{Cov}(\mathrm{IFD}_{\mathrm{cond}}, R_{\mathrm{token}}),
    \label{eq:varR-zac}
\end{multline}
where the outcome term vanishes by assumption. The following proposition formalizes that ISPO almost surely preserves the gradient even in this regime.

\paragraph{Proposition 2 (Gradient preservation under zero-advantage collapse).}
\emph{Suppose all $G$ rollouts share the same outcome, so $\sigma_{\mathrm{outcome}} = 0$. Let $\pi_\theta$ be a non-degenerate sampling policy in the sense that the joint distribution over $(T, A)$ is not a point mass. Then for any $\lambda_1 > 0$, $\Pr[\sigma_R > 0] = 1$.}

\noindent\emph{Proof sketch.}
Under non-degeneracy, the rollout trajectories $(T_i, A_i)$ are not all identical with probability one. Since $\mathrm{IFD}_{\mathrm{cond}}$ is a continuous, non-constant function of $(T, A)$ on the support of $\pi_\theta$, the realizations $\{\mathrm{IFD}_{\mathrm{cond}}^{(i)}\}_{i=1}^G$ are not all equal a.s., so $\mathrm{Var}(\mathrm{IFD}_{\mathrm{cond}}) > 0$ a.s. By~\eqref{eq:varR-zac}, $\sigma_R^2 \ge \lambda_1^2 \mathrm{Var}(\mathrm{IFD}_{\mathrm{cond}}) > 0$ a.s.\hfill$\square$
\label{prop:gradient-preserved}

This is the formal sense in which ISPO ``unlocks'' the zero-advantage regime: a batch that would be discarded by GRPO (or by dynamic sampling heuristics) still produces an informative update under ISPO.

\paragraph{Two caveats.} The non-degeneracy assumption is weakest late in training, when $\pi_\theta$ peaks and within-group rollouts grow similar: $\mathrm{Var}(\mathrm{IFD}_{\mathrm{cond}})$ then shrinks, so the guarantee, while almost sure, becomes numerically thin exactly where zero-advantage batches are most frequent; the token-level term supplies a second, independent variance source that mitigates this. Moreover, Proposition~\ref{prop:gradient-preserved} guarantees a \emph{non-zero}, not necessarily \emph{outcome-aligned}, gradient: in an all-wrong group the recovered signal ranks rollouts by thinking--answer dependence and critical-token entropy rather than by correctness, acting as exploration and calibration pressure rather than direct error correction. Since the per-rollout $\mathrm{IFD}_{\mathrm{cond}}$ is a single-sample Monte-Carlo estimate of the conditional KL, part of its variance is estimator noise; we keep $\lambda_1$ small so this refinement never overwhelms the outcome signal when present (Case~1). Appendix~\ref{sec:appendix-zac-recovery} shows that the recovered updates are nonetheless productive in practice.

\paragraph{Case 3: Late-stage hallucinated certainty.}
Late in training, $\pi_\theta$ becomes peaked at most tokens, including critical ones. For correct rollouts, this is desirable: the model is sharpening confidence in valid reasoning. For incorrect rollouts, however, the same peaking yields confidently wrong predictions, the failure mode that purely outcome-driven RL cannot diagnose, since the outcome reward treats all wrong trajectories identically. ISPO's incorrect-rollout reward~\eqref{eq:rtoken-neg} actively reverses this: the hinge term $\mathrm{pen}(H_t) = \max(0, \tau_H - H_t)$ becomes \emph{active} precisely when $H_t < \tau_H$, applying a corrective signal exactly where hallucinated certainty has set in. Across the three cases, ISPO is graceful where GRPO suffices, gradient-preserving where GRPO fails, and self-correcting where late-stage GRPO degrades.

\section{Additional Empirical Analysis}
\label{sec:appendix-empirical}

\subsection{Hyperparameter Sensitivity}
\label{sec:appendix-sensitivity}

% Hyperparameter sensitivity on Qwen2.5-Math-7B (Avg Pass@1 across 5 benchmarks).
\begin{table}[t]
\centering
\small
\setlength{\tabcolsep}{4pt}
\begin{tabular}{l c}
\toprule
\textbf{Configuration} & \textbf{Avg.\ Pass@1} \\
\midrule
\rowcolor{gray!15}\multicolumn{2}{c}{\textit{(a) Sequence-level weight $\lambda_1$}} \\
$\lambda_1 = 0$ (no IFD)         & 47.0 \\
$\lambda_1 = 0.1$                & 51.6 \\
$\lambda_1 = 0.5$ (default)      & \textbf{55.4} \\
$\lambda_1 = 1.0$                & 54.7 \\
$\lambda_1 = 2.0$                & 52.3 \\
\midrule
\rowcolor{gray!15}\multicolumn{2}{c}{\textit{(b) Token-level weight $\lambda_2$}} \\
$\lambda_2 = 0$ (no token reward)& 49.4 \\
$\lambda_2 = 0.05$               & 53.9 \\
$\lambda_2 = 0.10$ (default)     & \textbf{55.4} \\
$\lambda_2 = 0.30$               & 54.2 \\
\midrule
\rowcolor{gray!15}\multicolumn{2}{c}{\textit{(c) Critical-token fraction $k$\%}} \\
top-5\%                          & 53.6 \\
top-10\% (default)               & \textbf{55.4} \\
top-15\%                         & 54.8 \\
top-20\%                         & 53.3 \\
\midrule
\rowcolor{gray!15}\multicolumn{2}{c}{\textit{(d) Drift threshold $\tau_\Delta$}} \\
$\tau_\Delta = 0.1$              & 54.3 \\
$\tau_\Delta = 0.3$ (default)    & \textbf{55.4} \\
$\tau_\Delta = 0.5$              & 53.5 \\
\midrule
\rowcolor{gray!15}\multicolumn{2}{c}{\textit{(e) Hinge threshold $\tau_H$}} \\
$\tau_H = 0.1$                   & 54.0 \\
$\tau_H = 0.3$ (default)         & \textbf{55.4} \\
$\tau_H = 0.5$                   & 54.6 \\
\midrule
\rowcolor{gray!15}\multicolumn{2}{c}{\textit{(f) Penalty weights $(\alpha, \beta)$}} \\
$(\alpha,\beta) = (1.0, 0.0)$ (no hinge) & 51.7 \\
$(\alpha,\beta) = (1.0, 0.5)$            & 54.5 \\
$(\alpha,\beta) = (1.0, 1.0)$ (default)  & \textbf{55.4} \\
$(\alpha,\beta) = (1.0, 2.0)$            & 54.2 \\
\bottomrule
\end{tabular}
\caption{\textbf{Hyperparameter sensitivity on Qwen2.5-Math-7B.} Avg Pass@1 across AIME24/AIME25/AMC23/MATH-500/OlympiadBench. ISPO is robust within reasonable ranges of each hyperparameter; the default configuration (in bold) is used in all main-paper results. Performance degrades gracefully outside the defaults, with no setting collapsing below the GRPO baseline (44.1).}
\label{tab:sensitivity}
\end{table}

Table~\ref{tab:sensitivity} reports ISPO's sensitivity to each hyperparameter on Qwen2.5-Math-7B. Performance is robust within a wide neighborhood of the defaults: across all 22 configurations swept, ISPO never drops below 47.0 Avg Pass@1 (default 55.4), and always stays well above the vanilla GRPO baseline of 44.1. The default choice of $\lambda_1 = 0.5 > \lambda_2 = 0.1$ reflects that the sequence-level Conditional IFD is the primary driver of ISPO's gain: it carries the formal gradient-preservation guarantee of Proposition~\ref{prop:gradient-preserved} and acts on every rollout, while the token-level reward serves as a fine-grained refinement targeting hallucinated certainty (Failure Mode 2) at the $\sim$5--10\% of critical tokens. The 5$\times$ ratio between $\lambda_1$ and $\lambda_2$ also normalizes the raw scales of the two signals (the per-token IFD gap is typically $\sim 10\times$ smaller in magnitude than the token-level reward), so their actual contributions to $R(\tau)$ are comparable. The dominant failure modes are \emph{(i)} for $\lambda_1, \lambda_2$ too small, ISPO collapses toward GRPO (e.g., $\lambda_1=0$ yields 48.8); \emph{(ii)} for $\lambda_1, \lambda_2$ too large, the dense signals overpower the outcome reward (e.g., $\lambda_1=2.0$ yields 54.1); \emph{(iii)} for $k\%$ or $\tau_\Delta$ too low, the critical-token set becomes too sparse to deliver token-level gradient; too high, it dilutes the signal across non-critical tokens.

\subsection{Multi-seed Robustness}
\label{sec:appendix-robustness}

% Multi-seed robustness on Qwen2.5-Math-7B (3 seeds, Pass@1).
\begin{table}[t]
\centering
\small
\setlength{\tabcolsep}{4pt}
\begin{tabular}{l ccc}
\toprule
\textbf{Method} & \textbf{AIME24} & \textbf{MATH-500} & \textbf{OlympiadBench} \\
\midrule
Vanilla GRPO & 28.9$\pm$1.4 & 75.5$\pm$0.4 & 41.0$\pm$0.7 \\
Dr.\,GRPO           & 29.6$\pm$1.1 & 81.6$\pm$0.3 & 44.9$\pm$0.5 \\
PACR                 & 42.5$\pm$1.3 & 81.7$\pm$0.4 & 45.8$\pm$0.6 \\
\rowcolor{ispoblue!12}\textbf{ISPO (ours)}                      & \textbf{48.9}$\pm$1.1 & \textbf{83.6}$\pm$0.3 & \textbf{47.4}$\pm$0.5 \\
\bottomrule
\end{tabular}
\caption{\textbf{Multi-seed robustness} on Qwen2.5-Math-7B. Mean $\pm$ standard deviation of Pass@1 over 3 random seeds. ISPO's gains over each baseline exceed the inter-seed standard deviation by a wide margin, indicating that the improvements reported in Table~\ref{tab:main-math} are not artifacts of seed selection.}
\label{tab:multi-seed}
\end{table}

Table~\ref{tab:multi-seed} reports the mean and standard deviation of Pass@1 accuracy across three independent training runs (random seeds), measuring how sensitive each method is to seed selection. On AIME24, ISPO's gain over the strongest baseline (PACR, 42.5) is 6.4 percentage points, several times the per-method seed standard deviations (1.0--1.4). The same pattern holds on MATH-500 and OlympiadBench, indicating that the improvements reported in the main paper are not artifacts of seed selection. All other hyperparameters are held fixed at the defaults specified in Appendix~\ref{sec:appendix-sensitivity}.

\subsection{Conditional IFD Distribution over Training}
\label{sec:appendix-ifd-evolution}

\begin{figure*}[t]
  \centering
  \includegraphics[width=0.95\linewidth]{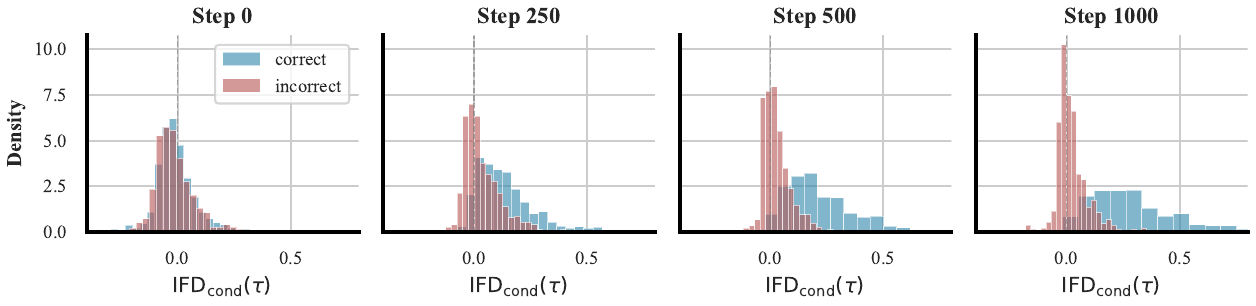}
  \caption{\textbf{Evolution of the Conditional IFD distribution across training} on Qwen2.5-Math-7B, split by rollout outcome. At step 0 (base policy), correct and incorrect distributions overlap around zero, indicating no learned thinking--answer dependence. As training progresses, the correct-rollout distribution shifts rightward (mean $\sim$0 $\to$ $\sim$0.3 with a heavy positive tail) while the incorrect-rollout distribution stays compact near zero, validating that the model learns to make $T$ informative for $A$ specifically on correct answers. The widening gap is precisely what enables signed-IFD to produce non-zero variance even in same-outcome groups.}
  \label{fig:ifd-evolution}
\end{figure*}

Figure~\ref{fig:ifd-evolution} traces the Conditional IFD distribution across four training stages on Qwen2.5-Math-7B. At initialization, correct and incorrect rollouts share a near-zero IFD distribution: the base policy's thinking trace has no learned dependence on the answer. As training progresses, the two distributions separate: correct rollouts develop a positive IFD tail (the model has learned to use $T$ to support the right answer), while incorrect rollouts stay near zero (the model has not learned to confidently rationalize wrong answers). This widening separation is what signed-IFD exploits to produce non-degenerate within-group variance even in zero-advantage batches, and it is the empirical mechanism behind Proposition~\ref{prop:gradient-preserved}.

\subsection{Zero-Advantage Batch Recovery}
\label{sec:appendix-zac-recovery}

\begin{figure}[t]
  \centering
  \includegraphics[width=0.85\columnwidth]{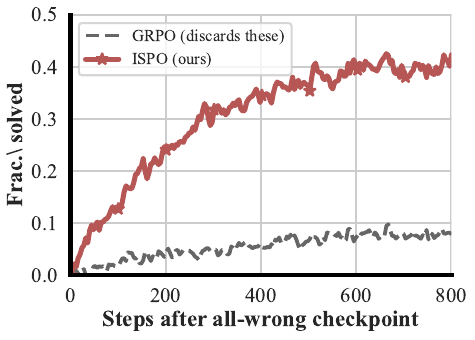}
  \caption{\textbf{Fate of zero-advantage prompts.} Among prompts whose rollout group is all-wrong ($\sigma_{\mathrm{outcome}} = 0$) at an early checkpoint (exactly the batches GRPO discards), the cumulative fraction subsequently solved (Pass@1) over continued training. GRPO improves on them only slowly and indirectly; ISPO keeps a non-zero gradient and converts a substantial fraction into solved prompts.}
  \label{fig:zac-recovery}
\end{figure}

To test whether the gradient ISPO recovers in zero-advantage batches (Case~2) is \emph{useful} rather than merely non-zero, we isolate the prompts whose rollout group is entirely incorrect at an early checkpoint and track the cumulative fraction the policy later solves. Figure~\ref{fig:zac-recovery} contrasts ISPO with GRPO on this set. Because GRPO produces no gradient on all-wrong groups, it improves on them only indirectly, through generalization from other prompts; ISPO's dense intrinsic signal keeps these prompts in the learning loop and converts a substantial fraction into solved prompts as training continues. This complements the almost-sure non-degeneracy guarantee of Proposition~\ref{prop:gradient-preserved} with a downstream-accuracy view, indicating that the recovered updates carry productive learning signal.

\end{document}